\documentclass[10pt,twocolumn,letterpaper]{article}

\usepackage[pagenumbers]{cvpr} 

\usepackage{graphicx}
\usepackage{amsmath}
\usepackage{amssymb}
\usepackage{booktabs}
\usepackage{enumitem}
\usepackage{graphicx,wrapfig,lipsum}

\usepackage{nicefrac}       
\usepackage{xcolor}
\usepackage[binary-units=true]{siunitx}
\usepackage{bm}
\usepackage{import}
\usepackage{comment}
\usepackage{subcaption}
\usepackage{stfloats}
\usepackage{transparent}

\DeclareMathOperator*{\argmin}{arg\,min}

\newcommand{\fz}{\bm{z}}
\newcommand{\fs}{\bm{s}}
\newcommand{\fp}{\bm{p}}
\newcommand{\fx}{\bm{x}}

\newcommand{\fw}{\bm{w}}

\newcommand{\rs}{\mathrm{s}}
\newcommand{\rp}{\mathrm{p}}
\newcommand{\rqq}{\mathrm{q}}
\newcommand{\norm}[1]{\left\|#1\right\|}

\newcommand\blfootnote[1]{%
  \begingroup
  \renewcommand\thefootnote{}\footnote{#1}%
  \addtocounter{footnote}{-1}%
  \endgroup
}

\usepackage[pagebackref,breaklinks,colorlinks]{hyperref}

\usepackage[capitalize]{cleveref}
\crefname{section}{Sec.}{Secs.}
\Crefname{section}{Section}{Sections}
\Crefname{table}{Table}{Tables}
\crefname{table}{Tab.}{Tabs.}

\def\cvprPaperID{***} 
\def\confName{CVPR}
\def\confYear{2022}

\begin{document}
\newcommand{\OUR}{SPAM}
\newcommand{\OURS}{SPAMs}
\newcommand{\OURSFULL}{Structured-implicit PArametric Models}

\title{SPAMs: Structured Implicit Parametric Models}

\author{
Pablo Palafox$^{1*}$~~~~~~
Nikolaos Sarafianos$^2$~~~~~~
Tony Tung$^2$~~~~~~
Angela Dai$^1$
\vspace{0.2cm} \\ 
$^1$Technical University of Munich~~~
$^2$Meta Reality Labs Research, Sausalito, USA
\vspace{0.2cm} \\ 
\href{https://pablopalafox.github.io/spams}{https://pablopalafox.github.io/spams}
}

\twocolumn[{
	\renewcommand\twocolumn[1][]{#1}%
	\maketitle
	\begin{center}
	    \vspace{-0.425cm}
		\fontsize{9pt}{11pt}\selectfont
        \def\svgwidth{\linewidth}
\begingroup%
  \makeatletter%
  \providecommand\color[2][]{%
    \errmessage{(Inkscape) Color is used for the text in Inkscape, but the package 'color.sty' is not loaded}%
    \renewcommand\color[2][]{}%
  }%
  \providecommand\transparent[1]{%
    \errmessage{(Inkscape) Transparency is used (non-zero) for the text in Inkscape, but the package 'transparent.sty' is not loaded}%
    \renewcommand\transparent[1]{}%
  }%
  \providecommand\rotatebox[2]{#2}%
  \newcommand*\fsize{\dimexpr\f@size pt\relax}%
  \newcommand*\lineheight[1]{\fontsize{\fsize}{#1\fsize}\selectfont}%
  \ifx\svgwidth\undefined%
    \setlength{\unitlength}{735.78413351bp}%
    \ifx\svgscale\undefined%
      \relax%
    \else%
      \setlength{\unitlength}{\unitlength * \real{\svgscale}}%
    \fi%
  \else%
    \setlength{\unitlength}{\svgwidth}%
  \fi%
  \global\let\svgwidth\undefined%
  \global\let\svgscale\undefined%
  \makeatother%
  \begin{picture}(1,0.42670294)%
    \lineheight{1}%
    \setlength\tabcolsep{0pt}%
    \put(0,0){\includegraphics[width=\unitlength,page=1]{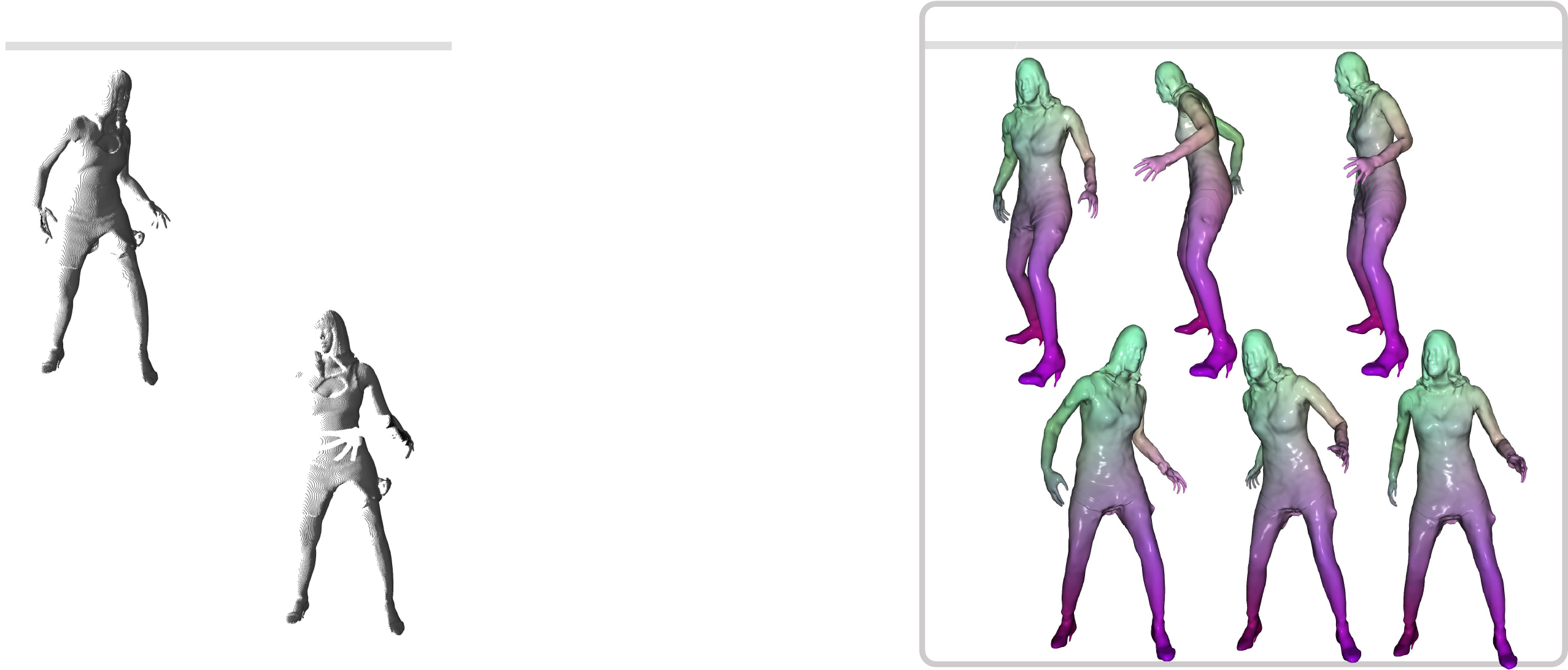}}%
    \put(0.61615758,0.24540244){\color[rgb]{0,0,0}\rotatebox{90}{\makebox(0,0)[lt]{\lineheight{1.25}\smash{\begin{tabular}[t]{l}Frontal Views\end{tabular}}}}}%
    \put(0.07619542,0.40567183){\color[rgb]{0,0,0}\makebox(0,0)[lt]{\lineheight{1.25}\smash{\begin{tabular}[t]{l}Input Depth Sequence\end{tabular}}}}%
    \put(0.69665547,0.40512034){\color[rgb]{0,0,0}\makebox(0,0)[lt]{\lineheight{1.25}\smash{\begin{tabular}[t]{l}Reconstruction and Tracking\end{tabular}}}}%
    \put(0.61615758,0.04927647){\color[rgb]{0,0,0}\rotatebox{90}{\makebox(0,0)[lt]{\lineheight{1.25}\smash{\begin{tabular}[t]{l}Lateral Views\end{tabular}}}}}%
    \put(0.05856961,0.16802487){\color[rgb]{0,0,0}\makebox(0,0)[lt]{\lineheight{1.25}\smash{\begin{tabular}[t]{l}$t_0$\end{tabular}}}}%
    \put(0.13652557,0.08947683){\color[rgb]{0,0,0}\makebox(0,0)[lt]{\lineheight{1.25}\smash{\begin{tabular}[t]{l}$t_1$\end{tabular}}}}%
    \put(0.21397448,0.01995462){\color[rgb]{0,0,0}\makebox(0,0)[lt]{\lineheight{1.25}\smash{\begin{tabular}[t]{l}$t_2$\end{tabular}}}}%
    \put(0.69846018,0.01499885){\color[rgb]{0,0,0}\makebox(0,0)[lt]{\lineheight{1.25}\smash{\begin{tabular}[t]{l}$t_0$\end{tabular}}}}%
    \put(0.82318327,0.01499885){\color[rgb]{0,0,0}\makebox(0,0)[lt]{\lineheight{1.25}\smash{\begin{tabular}[t]{l}$t_1$\end{tabular}}}}%
    \put(0.91593072,0.01499885){\color[rgb]{0,0,0}\makebox(0,0)[lt]{\lineheight{1.25}\smash{\begin{tabular}[t]{l}$t_2$\end{tabular}}}}%
    \put(0,0){\includegraphics[width=\unitlength,page=2]{teaser.pdf}}%
    \put(0.36534298,0.40809562){\color[rgb]{0,0,0}\makebox(0,0)[lt]{\lineheight{1.25}\smash{\begin{tabular}[t]{l}Part Prediction on Input\end{tabular}}}}%
    \put(0,0){\includegraphics[width=\unitlength,page=3]{teaser.pdf}}%
    \put(0.33966267,0.25390207){\color[rgb]{0,0,0}\makebox(0,0)[lt]{\lineheight{1.25}\smash{\begin{tabular}[t]{l}$t_0$\end{tabular}}}}%
    \put(0,0){\includegraphics[width=\unitlength,page=4]{teaser.pdf}}%
    \put(0.42289692,0.25325667){\color[rgb]{0,0,0}\makebox(0,0)[lt]{\lineheight{1.25}\smash{\begin{tabular}[t]{l}$t_1$\end{tabular}}}}%
    \put(0,0){\includegraphics[width=\unitlength,page=5]{teaser.pdf}}%
    \put(0.47840154,0.25261127){\color[rgb]{0,0,0}\makebox(0,0)[lt]{\lineheight{1.25}\smash{\begin{tabular}[t]{l}$t_2$\end{tabular}}}}%
    \put(0,0){\includegraphics[width=\unitlength,page=6]{teaser.pdf}}%
    \put(0.410143,0.16827706){\color[rgb]{0,0,0}\makebox(0,0)[lt]{\lineheight{1.25}\smash{\begin{tabular}[t]{l}Part Space\end{tabular}}}}%
  \end{picture}%
\endgroup%

	    \vspace{-0.5cm}
		\captionof{figure}{
		    We propose to represent deforming shapes with a structural decomposition into part-based disentangled spaces characterizing of shape and pose, as \OURSFULL{} (\OURS{}).
		    \OURS{} learn optimizable local shape and pose spaces, which we can traverse at test time to fit to depth sequence observations of an unseen deforming object.
		    Our structured part decompositions facilitate low-dimensional coarse motion correspondence through part-to-part correlation, which guides a robust joint optimization over the part-based shape and pose spaces for globally consistent, accurate tracking under complex motion sequences.
		}
		\label{fig:teaser}
		\vspace{-0.075cm}
	\end{center}
}]



\begin{abstract}
Parametric 3D models have formed a fundamental role in modeling deformable objects, such as human bodies, faces, and hands; however, the construction of such parametric models requires significant manual intervention and domain expertise.
Recently, neural implicit 3D representations have shown great expressibility in capturing 3D shape geometry.
We observe that deformable object motion is often semantically structured, and thus propose to learn \OURSFULL{} (\OURS) as a deformable object representation that structurally decomposes non-rigid object motion into part-based disentangled representations of shape and pose, with each being represented by deep implicit functions.
This enables a structured characterization of object movement, with part decomposition characterizing a lower-dimensional space in which we can establish coarse motion correspondence.
In particular, we can leverage the part decompositions at test time to fit to new depth sequences of unobserved shapes, by establishing part correspondences between the input observation and our learned part spaces; this guides a robust joint optimization between the shape and pose of all parts, even under dramatic motion sequences.
Experiments demonstrate that our part-aware shape and pose understanding lead to state-of-the-art performance in reconstruction and tracking of depth sequences of complex deforming object motion.
We plan to release models to the public.
\end{abstract}
\section{Introduction}~\label{sec:intro}

\blfootnote{*This work was conducted during an internship at Meta RL Research.}
Understanding non-rigidly deforming shapes is essential for real-world perception, as we live in a 4D world where humans, animals, and many other 3D objects move in a non-rigid fashion.
Dynamic tracking and reconstruction remains a notable challenge, and while significant advances have been made in 4D reconstruction and tracking, they often require complex multi-view setups~\cite{bozic2020neural} or build on a domain-specific, fixed-topology template~\cite{pons2015dyna, alldieck2019learning, pons2017clothcap, bhatnagar2019multi}.
In the latter scenario, parametric 3D models in particular have made notable impact in modeling domain-specific deformable 3D objects, such as for human bodies~\cite{anguelov2005scape, loper2015smpl, joo2018total}, faces~\cite{li2017learning_flame, paysan20093d}, hands~\cite{MANO:SIGGRAPHASIA:2017}, and animals~\cite{Zuffi_CVPR_2017}.
However, such parametric 3D models require a complex construction process involving domain-specific knowledge and manual efforts, while remaining limited in expressability of local shape details.

Recently, advances in learned continuous implicit representations for modeling 3D shapes have shown impressive representation power for capturing effective static 3D shape geometry at relatively high resolutions~\cite{chen2019learning, genova2019learning, mescheder2019occupancyNet, michalkiewicz2019deep, park2019deepsdf, chibane2020implicit, ramon2021h3d}.
Such approaches have also been extended to represent 4D reconstruction of dynamic objects by efficiently disentangling learned implicit spaces representing shape and dynamic movement~\cite{palafox2021npms, niemeyer2019occupancyFlow}.
This has proven to be a very promising direction, but these approaches characterize objects as a whole, whereas we observe that the 4D motion of an object typically maintains a strong structured correlation on a lower-level part basis.

Thus, we propose \OURSFULL{} (\OURS{}), which learn a structured, part-based, disentangled representation of deformable 3D objects. 
Given a set of observations of various shape identities in different poses (including a canonical pose) with coarse part annotations, we learn part-based latent spaces characterizing each part's geometry and motion.
Note that we do not require comprehensive surface correspondence throughout the dataset, nor complex domain-specific knowledge (e.g., skeleton, kinematic chain).
We leverage continuous implicit function representations for each part's geometry, represented as a signed distance field in its canonical space, and pose, represented as a local deformation relative to the canonical space.

At test time, we traverse the learned latent part spaces to fit to new depth sequences.
Crucially, our part-based representation allows leveraging predicted part segmentation of the new observation to establish global correspondences with our part-based latent representations.
By establishing correspondence through our part priors, we can robustly track sequences with significant motion changes by discovering high-level part correspondence and leveraging it to guide our joint optimization over part-based shape and pose.
Experiments on non-rigid tracking and reconstruction of single-camera depth sequences of humans from the RenderPeople dataset~\cite{renderppl} show that the part-aware reasoning of our \OURS{} can outperform the state of the art by an order of magnitude on reconstruction (Chamfer distance) and by $43\%$ on tracking (3D End-Point-Error). 
%
In summary, we present the following contributions:
\begin{itemize}[leftmargin=*]
    \item We learn a part-based disentanglement of shape and pose, capturing local characteristics of deformable 3D objects in latent spaces representing shape and pose of each part.
    \item Our learned, optimizable spaces enable part-based reasoning to guide joint optimization over parts to fit unseen test sequences.
    By establishing high-level part correspondence between a new observation and our learned part spaces, we can robustly guide a joint optimization over part geometry and pose, resulting in a more globally consistent non-rigid reconstruction and tracking. 
\end{itemize}

\section{Related Work}

\noindent\textbf{Parametric and Neural Parametric Models. }
Parametric body models enable the representation of the human body variations with a limited number of parameters (\ie a low-dimension descriptor). For example, SMPL~\cite{loper2015smpl} is a parametric model widely-used to describe the body shape and pose with deformation blend shapes learned from a dataset of diverse 3D body scans.
Parametric body models have enabled research work for modeling soft tissue~\cite{pons2015dyna} and clothing~\cite{ma2020learning, alldieck2019learning,tiwari2020sizer}. 
However parametric body models such as SMPL~\cite{loper2015smpl} or GHUM~\cite{xu2020ghum} employ deformations based on vertex-based skinning models which have limited resolution and cannot represent non-linear surface deformations of clothed bodies (\eg, wrinkles).
To alleviate these limitations, Neural Parametric Models~\cite{palafox2021npms} learn to disentangle 4D dynamics into latent-space representations of shape and pose using implicit functions and can fit to new observations by optimizing over the pose and shape codes. However, they treat the human body as a single entity, which results in somewhat unrealistic motions. 
Genova \etal~\cite{genova2020ldif, genova2019learning} addressed this problem by introducing LDIF, a 3D representation that implicitly describes a shape as the sum of local 3D functions.
LDIF outputs a structured decomposition into shape elements by encoding 3D points within each shape using PointNet~\cite{fan2017point, qi2017pointnet, qi2017pointnetplus} and hence enable shape encoding such as human bodies in local regions arranged in a global structure.
The 3D decompositions of LDIF tend towards temporal consistency, but as tracking is not explicitly considered, surface tracking tends to become inconsistent in more challenging motion scenarios.
In contrast, we learn a semantically-driven part decomposition which guides joint part-based shape and pose optimization that results in robust, consistent tracking over dynamic sequences.\\

\noindent\textbf{Continuous Implicit Deformable Representations.}
Implicit-based representations for humans or clothing has been an active topic of research for the past few years~\cite{chibane20ifnet, chibane2020ndf,park2019deepsdf,LEAP_CVPR}.
Several recent works have focused on learning identity-specific implicit representations to animate clothed people \cite{tiwari21neuralgif, LEAP_CVPR, Saito_CVPR2021, jeruzalski2020nilbs,deng2020nasa, corona2021smplicit}.
Neural-GIF~\cite{tiwari21neuralgif} introduced a framework to animate clothed people from scans as a function of pose directly, without the need of registration.
SCANimate~\cite{Saito_CVPR2021} and LEAP~\cite{LEAP_CVPR} learn a pose or shape representation of the human body surface.
While promising, these methods learn subject or outfit-specific models, and thus lack general characterization of deformable objects. 
Given a posed but unclothed body model, POP~\cite{POP_ICCV2021} represents a clothed human with a set of points in order to create pose-dependent clothing animations. 
New animations can then be created from an unseen scan of a person in clothing.
Bhatnagar \etal~\cite{bhatnagar2020ipnet} proposed IP-Net, a method to combine learned implicit functions and traditional parametric models to produce controllable models of humans.
IP-Net predicts correspondences to SMPL and leverages a double-layered surface to represent inner and outer surfaces to better represent a clothed body. 
Our proposed approach also builds on the expressability of learned implicit representations, and we propose to learn from a dataset without strong requirements on surface correspondence annotations to construct a general, semantically-driven decomposition that provides strong high-level guidance for robust pose tracking under challenging motion scenarios.

\begin{figure*}[t]
    \centering
    \fontsize{9pt}{11pt}\selectfont
    \def\svgwidth{0.95\linewidth}
    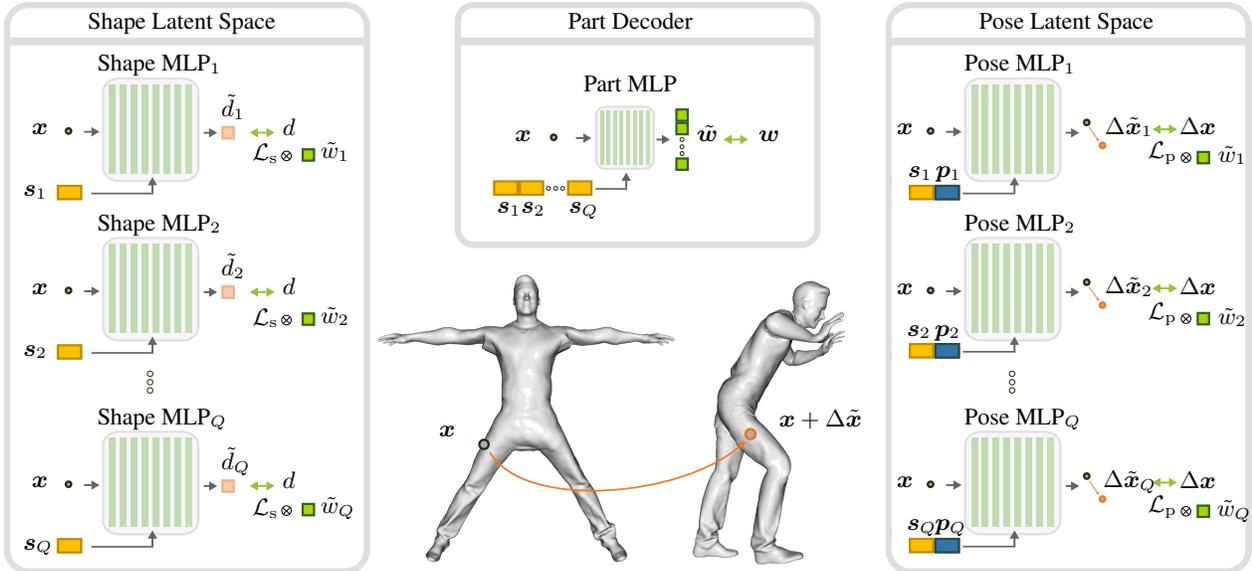
    \vspace{-0.2cm}
    \caption{
        \textbf{Structured Implicit Parametric Model Construction}. 
        \OURS{} learn a structured decomposition of shape and pose, where local shape codes $\{\fs_{q}\}$ can be decoded to represent local part geometry, and both $\{\fs_{q}\}$ and a part decoder inform local pose movement.
        To construct a \OUR{}, we first learn structural decomposition in the canonical space as a part decoder that predicts part classes, conditioned on all local shape codes $\{\fs_{q}\}$; we use this learned part decomposition to guide local (shape and pose) MLPs into focusing on their corresponding space partitioning.
        We learn a structured, latent shape space in the canonical space by conditioning a set of local shape MLPs on local shape codes $\{\fs_{q}\}$ assigned to every identity and part $q$.
        We learn local deformation fields around the canonically-posed shape with local pose MLPs conditioned on a local, latent shape and pose codes $\fs_{q}$ and $\fp_{q}$ to predict $\Delta \fx$ for  $\fx$ in the canonical space.
    }
    \label{fig:part_aware_latent_space_learning}
\end{figure*}

\section{Structured Implicit Parametric Models}

We introduce \OURSFULL{} (\OURS{}), a learned approach to build part-aware, implicit, parametric 3D models from a dataset of posed identities of a given object class.
We do not require the dataset to have surface correspondence between each instance nor annotations for physical domain-specific properties such as the skeleton or the kinematic chain.
Our \OURS{} structurally decompose non-rigid object motion into part-based disentangled implicit spaces representing each part's shape and pose.

\OURS{} consist of (1) learned latent spaces that characterize part geometries in the canonical shape space, (2) a part decoder conditional on the part shapes that guides a part-based structural partitioning of the canonical space, and (3) structured, latent pose spaces conditional on both part shape and pose.
At test time, \OURS{} not only allow for the joint optimization over the learned, local spaces of shape and pose to fit to a new observation, but, crucially, our part-based formulation enables establishing high-level correspondence between predicted parts and the part-based latent representations.
This part correspondence enables robust global optimization over the part shape and pose spaces, resulting in robust, consistent non-rigid reconstruction and tracking of unseen sequences of deforming objects.

\subsection{Overview}
Given a dataset of shape identities from the same class category in different poses, with coarse part segmentation in the canonical space, our goal is to learn a part-based parametric model that structurally disentangles shape and pose.
We leverage this structured part decomposition to fit to unseen depth sequences of new identities, where our part-based representation provides a lower-dimensional space to establish semantic part correspondences that provide strong guidance for our part-based shape and pose fitting under complex motions.

Our \OURS{} characterize disentangled shape and pose as sets of local, part-based shape spaces over shape codes $\{\fs_{q}\}$, and their corresponding local pose spaces over $\{\fp_{q}\}$.
To obtain the structured partitioning of these local spaces, we first learn a part decoder that predicts a segmentation of the canonical space into a set of local parts (Sec.~\ref{sec:part_decoder}).
Given the learned space partitioning, we can then construct the structured shape space as a set of local implicit geometric representations (Sec.~\ref{sec:learned_shape_space}).
We then build the structured pose space as a set of local implicit deformation fields that characterize the motion of each local shape (Sec.~\ref{sec:learned_pose_space}).
Finally, to fit to new depth sequences of unobserved identities at test time, we leverage part correspondence to the set of shape and pose spaces to obtain a robust, joint optimization that accurately represents the observed shape (Sec.~\ref{sec:test_time_optim}).

\subsection{Part Decoder} \label{sec:part_decoder}
In order to establish a structural space partitioning, we learn a part decoder that segments the canonical space and guides the learning of the part-based shape and pose latent spaces.
The part decoder $f_{\theta_{\rqq}}$ is characterized by an MLP that predicts part confidences for a query point $\fx \in \mathbb{R}^3$ in canonical space.
We implement our part decoder in an auto-decoder fashion~\cite{park2019deepsdf} and condition on the concatenation of $D_{\rs}$-dimensional local latent shape codes $\fs_{\rqq}$ (Sec.~\ref{sec:learned_shape_space}).
Let $Q$ denote the number of parts, leading to a $Q D_{\rs}$-dimensional concatenation of local shape codes.
Formally, we have:
\begin{equation}\small
    f_{\theta_{\rqq}} : \mathbb{R}^{Q D_{\rs}} \times \mathbb{R}^3 \rightarrow  \mathbb{R}^Q, \quad \left( [\fs_{\rqq}], \fx \right) \mapsto f_{\theta_{\rqq}} \left( [\fs_{\rqq}], \fx \right) = \tilde{\fw}.
\end{equation}
where $[\fs_{\rqq}]$ denotes concatenation of all $Q$ local shape codes~$\{\fs_{\rqq}\}$.
We train $f_{\theta_{\rqq}}$ with a binary cross entropy loss between predicted and ground truth part labels as one-hot vectors.

We consider six coarse parts: head, torso, right arm, right leg, left arm and left leg (see Fig.~\ref{fig:teaser}-center).
A point $\fx \in \mathbb{R}^3$ in canonical space can belong to any of these part classes, where the $Q$-dimensional $\tilde{\fw}$ denotes the likelihoods of $\fx$ belonging to each respective part.
In particular, we allow points near the boundary between two parts to belong to both parts at the same time, by indicating both parts to be one in the ground truth part vector.
This allows regularization of boundary regions to ensure smooth transitions between implicit functions for shape and pose latent spaces.

Note that $f_{\theta_{\rqq}}$ predicts part probabilities for all points $\fx \in \mathbb{R}^3$ in canonical space, not only surface points.
This strategy provided more accurate part predictions for off-surface locations during our test-time optimization (Sec.~\ref{sec:test_time_optim}).

\noindent\textbf{Implementation details.} 
We use a 6-layer SIREN \cite{sitzmann2020implicit} MLP with a hidden dimension of $256$ and a frequency $\omega = 30$ in the sinus activation (following \cite{sitzmann2020implicit}).
As shown by \cite{chan2021pi}, conditioning-by-concatenation is sub-optimal for implicit neural representations with period activations, and therefore we employ their proposed FiLM conditioning, where a mapping network takes in a latent code $\fz$ and outputs frequencies and phase to condition each layer of the SIREN MLP.
The mapping network is a 4-layer LeakyReLU MLP with hidden dimension of 128.
We use the Adam optimizer~\cite{kingma2014adam} and a learning rate of \SI{1e-4}{} for both the decoder and the mapping network.

\subsection{Structured Shape Space} \label{sec:learned_shape_space}
Our multi-part shape space is learned by a dictionary of $Q$ local MLPs, each learning to represent a local shape part in its canonical pose, characterizing its geometry as the zero iso-surface decision boundary of a signed distance field.
The structural decomposition of the $Q$ local MLPs is guided by the part decomposition predicted by the part decoder $f_{\theta_{\rqq}}$.
To extract the complete shape from our structured shape space, we  query all $Q$ shape MLPs for every point in a 3D grid, average out their SDF contributions based on predicted part confidence for the given query point, and finally use Marching Cubes~\cite{lorensen1987marching} to extract a mesh.

Each shape MLP is trained in auto-decoder fashion~\cite{park2019deepsdf}, similar to the part decoder.
We directly optimize over a latent code $\fs_{\rqq}$, which is particular to the shape latent space of its assigned part $q$. 
Each part $q$ of a canonically-posed shape identity $i$ in the training set is then encoded in a $D_{\rs}$-dimensional latent shape code $\fs_{\rqq}^i$.
In turn, each shape MLP $f_{\theta_{\rs}^{q}}$ learns to map an input point $\fx \in \mathbb{R}^3$ in the canonical space, conditioned on the local shape code $\fs_{\rqq}^i$, to a predicted SDF value $\tilde{d}_{\rqq}$:
\begin{equation}
    f_{\theta_{\rs}^{\rqq}} : \mathbb{R}^{D_{\rs}} \times \mathbb{R}^3 \rightarrow \mathbb{R}, \quad \left( \fs_{\rqq}^i, \fx \right) \mapsto f_{\theta_{\rs}^{\rqq}} \left( \fs_{\rqq}^i, \fx \right) = \tilde{d}_{\rqq}.
\end{equation}

As train data typically do not contain watertight meshes, we train directly on oriented point clouds sampled from the (potentially incomplete) train meshes, following~\cite{gropp2020implicit}.
This is accomplished by solving for an Eikonal boundary value problem that constrains the norm of spatial gradients of the SDF to be 1 almost everywhere \cite{gropp2020implicit, sitzmann2020implicit}.
In practice, we use the Eikonal loss as presented in \cite{sitzmann2020implicit}.
As training data, we only require the surface points with associated normals for each identity, and randomly sampled coordinates in the unit cube.
For every identity in a batch, we sample $N_{\rs}$ points, half of which are surface points subsampled from the given identity, and the remaining half as coordinates randomly sampled in space.
Then for every part $q$, we train the corresponding shape MLP by minimizing the following reconstruction energy over the $S$ canonically-posed shape identities of the dataset with respect to the local shape codes $\{ \fs_{\rqq}^i \}_{i=1}^S$ and the set of shape MLP weights $\theta_{\rs}^{\rqq}$:
\begin{equation}\small
    \argmin_{\theta_{\rs}^{\rqq}, \{ \fs_{\rqq}^i \}_{i=1}^S} \sum_{q=1}^Q \sum_{i=1}^S \Big(  \sum_{k=1}^{N_{\rs}} w_{\rqq}^{i,k} \mathcal{L}_{\rs}(f_{\theta_{\rs}^{\rqq}} (\fs_{\rqq}^i, \fx_i^k), d_i^k) + \mathcal{L}_{\rm{r}} \Big).
\end{equation}
Here $\mathcal{L}_s$ is the Eikonal loss version proposed in SIREN \cite{sitzmann2020implicit}, enforcing that (1) SDF predictions for surface points are 0, (2) groundtruth surface normals match the estimated normals (computed as the spatial gradient of the SDF function at a given position), (3) the norm of the SDF gradient is 1 almost everywhere and (4) off-surface points do not have SDF values close to 0.
We refer to \cite{sitzmann2020implicit} for a more detailed explanation.
Additionally, we enforce a zero-mean multivariate-Gaussian distribution with spherical covariance $\sigma_{\rs}$ over the latent shape codes, as was proposed in \cite{park2019deepsdf}: $\mathcal{L}_{\rm{r}} = \norm{\fs_{\rqq}^i}^2_2/\sigma_{\rs}^2$.

Importantly, the contribution to the loss of each point is weighted by its predicted part confidence $w_{\rqq}^{i,k}$ from $f_{\theta_{\rqq}}$, enabling each local shape MLP to focus on its respective local region.
That is, if for point $\fx$, $w_{\rqq}^{i,k}$ for a class $q$ is close to 1, then $\fx$ likely belongs to $q$.
Then gradients from the contribution of point $\fx$ will back-propagate with high weight to the shape MLP of part $q$, $\theta_{\rs}^{\rqq}$.
On the contrary, if $w_{\rqq}^{i,k}$ is near 0, then $\fx$ likely does not belong to $q$ and $f_{\theta_{\rqq}}$ will not learn to generate geometry at $\fx$.
%

\noindent\textbf{Implementation details.} 
Each shape latent space is implemented as a 6-layer $\mathrm{SIREN}$ \cite{sitzmann2020implicit} MLP with a hidden dimension of $256$ and a frequency $\omega = 30$.
We use $128$-dimensional shape latent codes for each part ($D_{\rs} = 128$).
Simliar to our part decoder, we employ FiLM conditioning \cite{chan2021pi} to condition the SDF predictions on the local latent shape code, with the mapping network implemented as a 4-layer LeakyReLU MLP with 128 units per layer.
We use the Adam optimizer and learning rates of \SI{1e-4}{} and \SI{1e-3}{} for the local shape MLPs $f_{\theta_{\rs}^{\rqq}}$ and the local shape codes $\{ \fs_{\rqq}^i \}_{i=1}^{S}$, respectively.
The latent shape codes are initialized randomly from $\mathcal{N}(0,\,0.01^{2})$.
 
\subsection{Structured Pose Space} \label{sec:learned_pose_space}
Similar to our structured shape latent space, our structured pose space is learned by a dictionary of $Q$ local MLPs, each optimized to represent a local deformation field that maps query points $\fx$ in the canonical space of an identity $i$ to a deformed space $j$, by predicting a flow vector $\Delta \tilde{\fx}_{\rqq}$.
This prediction is conditional on both a \mbox{$D_{\rp}$-dimensional} latent pose code $\fp_{\rqq}^j$ as well as on the latent shape code $\fs_{\rqq}^i$ of the corresponding part $q$, as pose deformations change with respect to shape.
Deformation fields are only defined for a thin layer around the shape surface, since flow vectors become less informative when further away from the surface.
Formally, we have:
\begin{align*}
    f_{\theta_{p}^{\rqq}} : \mathbb{R}^{D_{\rs}} \times \mathbb{R}^{D_{\rp}} \times \mathbb{R}^3 &\rightarrow \mathbb{R}^3, \\
    \left( \fs_{\rqq}^i, \fp_{\rqq}^j, \fx \right) &\mapsto f_{\theta_{p}^{\rqq}} \left( \fs_{\rqq}^i, \fp_{\rqq}^j, \fx \right) = \Delta \tilde{\fx}_{\rqq}.
\end{align*}
Our local pose MLPs are trained with up to $P$ deformed instances of an identity; note that this does not require different identities to appear in the same pose.

To train these local pose spaces, we sample dense, near-surface correspondences between the canonical and posed frame, which amounts to sampling the canonical and posed raw meshes at the same barycentric coordinates, similar to \cite{palafox2021npms}.
Learning one of the local pose spaces $\theta_{\rp}^{\rqq}$ amounts to minimizing the following energy term over all $P$ deformation fields with respect to the individual (and local) pose codes $\{ \fp_{\rqq}^j \}_{j=1}^P$ and pose MLP weights $\theta_{\rp}^{\rqq}$:
\begin{equation}
    \argmin_{\theta_{\rp}^{\rqq}, \{ \fp_{\rqq}^j \}_{j=1}^P}
    \sum_{\substack{j=1 \\ i(j)}}^P \Big(  \sum_{q=1}^Q \sum_{k=1}^{N_{\rp}} w_{\rqq}^{i,k} \mathcal{L}_{\rp} + \mathcal{L}_{\rm{r}} \Big),
\end{equation}
where $i(j)$ is a mapping from the index $j$ of a posed shape to the corresponding index $i$ of its canonical shape and $\mathcal{L}_{\rp}$ is an $\ell_2$ loss between predicted and ground truth flow vectors:
\begin{equation}
    \mathcal{L}_{\rp} := \mathcal{L}_{\rp}(f_{\theta_{\rp}^{\rqq}} (\fs_{\rqq}^i, \fp_{\rqq}^j, \fx_i^k), \Delta \fx_{ij}^k).
\end{equation}

Similar to the structured local shape spaces, we encourage corresponding structural decomposition of the local part spaces by leveraging the predicted part confidences $w_{\rqq}^{i,k}$ from  $f_{\theta_{\rqq}}$.
We also enforce an analogous zero-mean multivariate-Gaussian distribution over the latent pose codes with $\mathcal{L}_{\rm{r}}$.
While learning these pose latent spaces, we do not optimize over the local shape codes.

\noindent\textbf{Implementation details.}
We implement our structured pose latent spaces as 4-layer SIREN MLPs with hidden dimension equal to $256$, $\omega = 15$, and set $D_{\rp} = 64$ as the dimension of our local pose codes.
We employ the same training scheme as used for training the shape spaces.

\subsection{Test-time Optimization} \label{sec:test_time_optim}
Our structured latent representations of shape and pose can be optimized over at test time to fit a \OUR{} to accurately reconstruct and track an input sequence of $L$ depth maps.
This is achieved by solving for the set of $Q$ local latent shape codes, denoted as $[\tilde{\fs}]$, and the $Q \times L$ latent pose codes ($Q$ pose codes per frame), $\{ [\tilde{\fp}_j] \}_{j=1}^L$, that best explain the input sequence.

Each depth map in the sequence is interpreted as a \mbox{$256^3$-SDF} grid of its back-projected values.
A volumetric mask $M_{\mathrm{o}}$ is also extracted to mask out regions that are further than \SI{0.02}{} (in normalized units) from the observed surface.
We additionally predict part labels for every point in the input depth map using PointNet++ \cite{qi2017pointnet++}, which we pre-train to predict part labels for the $Q$ parts.

Prior to optimization, we initialize our local shape and pose codes.
To initialize each local shape code $\fs_{\rqq}$, we use the average optimized train code for each part $q$.
To initialize local pose codes, we leverage a learned pose encoder that maps the input depth map to a latent code.
While we find that our structured part representations of shape and pose can robustly track from a random pose initialization of all codes from $\mathcal{N}(0,\,0.01^{2})$, we can obtain improved pose tracking by a learned encoder initialization (c.f. Sec.~\ref{sec:results}).

Given the initial shape code estimates, we can extract an initial canonical shape by querying our structured shape MLPs on a 3D grid, and extracting the iso-surface with  Marching Cubes~\cite{lorensen1987marching}.
We use the initial canonical shape to inform sampling near the surface, and sample \mbox{$N_t = \SI{500}{}$k} points, \mbox{$\{ \fx_k \}_{k=1}^{N_t}$}, around this initial estimate of the canonical shape; during optimization, for each frame in a mini-batch, we sub-sample \mbox{$N_b = 20$}k points out of the available $N_t$ to minimize the following equation:
\begin{equation}\small
    [\tilde{\fs}_{\rqq}], \{ [\tilde{\fp}_{\rqq}^j] \}_{j=1}^L = \argmin_{[\fs_{\rqq}], \{ [\fp_{\rqq}^j] \}_{j=1}^L} \sum_{j=1}^L \sum_{\forall \fx_k} \mathcal{L}_{\mathrm{r}} + \mathcal{L}_{\mathrm{c}} + \mathcal{L}_{\mathrm{t}} + \mathcal{L}_{\mathrm{icp}}.
    \label{eq:test_time_optim}
\end{equation}
$\mathcal{L}_{\mathrm{c}}$ enforces shape and pose code regularization as in training, and $\mathcal{L}_{\mathrm{t}}$ enforces temporal regularization  between the current frame $j$ and its neighboring frames (for more detail, we refer to the supplementary material).

\subsubsection{Structurally-Guided Shape {\&} Pose Optimization}
\begin{figure}[t]
    \centering
    \fontsize{9pt}{11pt}\selectfont
    \def\svgwidth{0.75\linewidth}
    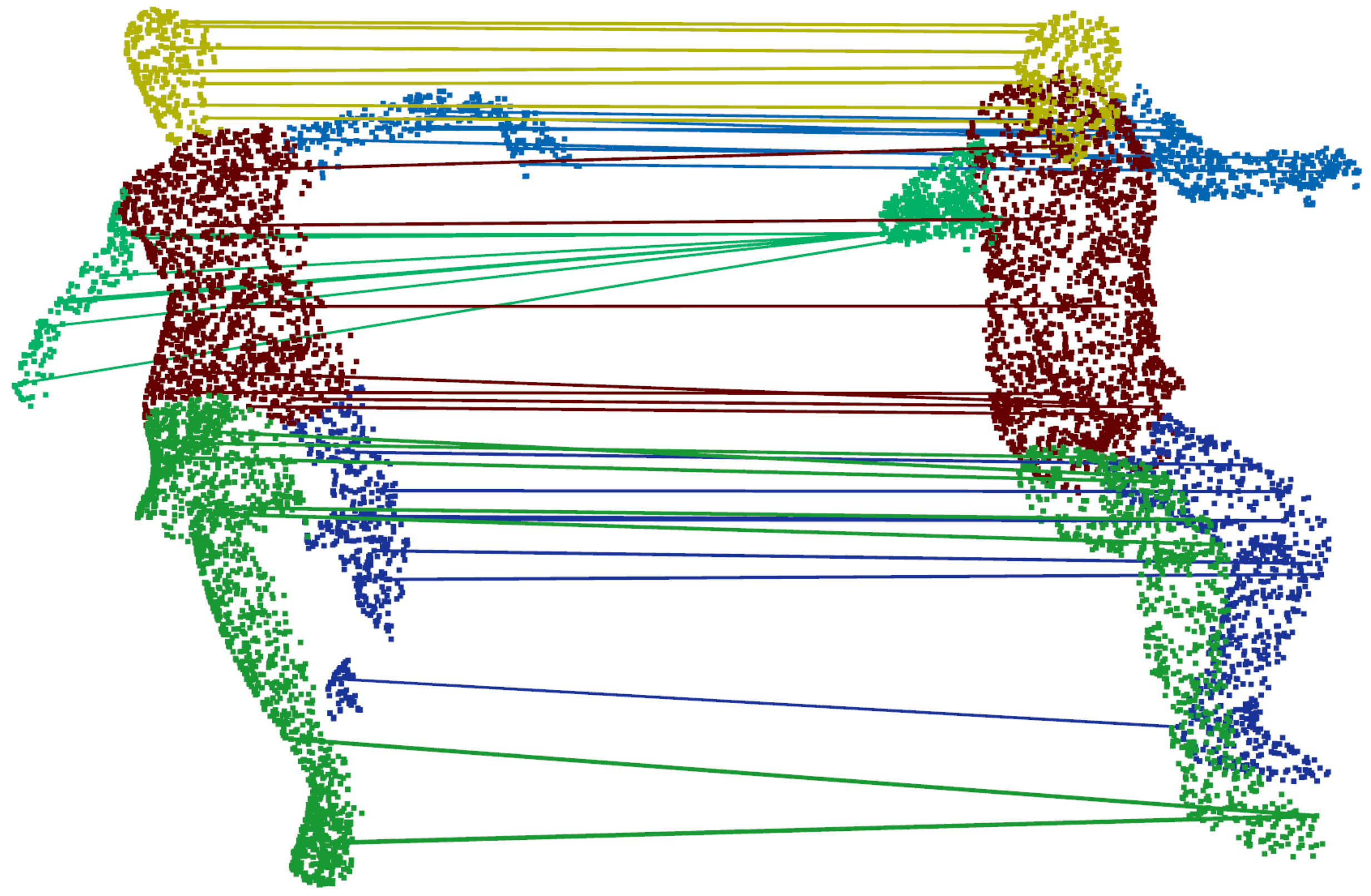
    \vspace{-0.25cm}
    \caption{
        Our part-aware $\mathcal{L}_{\mathrm{icp}}$ can establish high-level correspondences between input points (left) and posed estimates (right), even under difficult motions with poor initial pose estimates (e.g., right arm in green), thus guiding optimization to accurate tracking.
    }
    \label{fig:corresp}
\end{figure}
In order to inform our reconstruction losses $\mathcal{L}_{\mathrm{r}}$ and $\mathcal{L}_{\mathrm{icp}}$, we employ structural correspondences between parts predicted in the observed depth views as well as the part decomposition from our latent shape and pose spaces (see Fig.~\ref{fig:corresp}).

We use a clamped $\ell_1$ \cite{duan2020curriculum} reconstruction loss $\mathcal{L}_{\mathrm{r}}$:
\begin{equation}
    \mathcal{L}_{\mathrm{r}} = \sum_{q=1}^Q M \mathcal{L}_{\rs} \Big( f_{\theta_{\rs}^{\rqq}}(\fs_{\rqq}, \fx_k), \left[ \fx_k + f_{\theta_{\rp}^{\rqq}}(\fs_{\rqq}, \fp_{\rqq}^j, \fx_k) \right]_{\mathrm{sdf}} \Big),
\end{equation}
where $\left[ \cdot \right]_{\mathrm{sdf}}$ denotes trilinear interpolation of the SDF grid and $M = M^{\rqq}_{\rm{part}} M_{\mathrm{o}}$ leverages part and occlusion information to inform the reconstruction.
$M_{\mathrm{o}}$ denotes a mask of unoccluded regions, as defined in Sec.~\ref{sec:test_time_optim}.
$M_{\rm{part}}$ represents a grid of part label predictions, as given by the part decoder.
We compute $M_{\rm{part}}$ by randomly sampling points in the canonical space along with their predicted part labels from $f_{\theta_{\rqq}}$, warping these points to each frame by $f_{\theta_{\rp}^{\rqq}}$, and finally querying for every grid location the nearest point from the warped points to obtain the part label for the voxel.
This enables focusing the reconstruction locally for each estimated part geometry and pose.

Finally, we employ a part-guided ICP-inspired loss $\mathcal{L}_{\mathrm{icp}}$, which plays a key role in ensuring robustness to pose code initialization.
That is, we establish part-driven correspondences between each estimated part $q$ from the predicted input depth map part labels and the part decoder predictions.
To this end, every $I_{\rm{resample}}$ iterations we consider the canonically-posed shape from the current state of the shape codes, re-sample a new set of \mbox{$N_t = \SI{500}{}$k} points around the mesh and keep those within a distance $\epsilon_{\mathrm{icp}}$ from the implicitly represented surface.
We use our part decoder (Sec. \ref{sec:part_decoder}) to estimate part labels for these canonical points, and then warp them into a posed frame $j$ using our pose decoders.
Then for every point in the input depth map, and given its predicted part $q$ (obtained by PointNet++), we find its nearest neighbor in the warped set of points belonging to $q$ based on $f_{\theta_{\rqq}}$ to establish correspondences.
Crucially, this provides robust correspondences in challenging motion scenarios where pose initialization may be notably misaligned, as demonstrated in Sec.~\ref{sec:results}.

\section{Experiments}
\label{sec:results}

We evaluate \OURS{} on the task of model fitting to depth sequences (Sec.~\ref{sec:model_fitting_experiments}), and analyze the effect of our structured, part-based fitting in Sec.~\ref{subsec:ablations}.
%

\noindent\textbf{Datasets.}
We train and evaluate on the public RenderPeople dataset~\cite{renderppl}, which contains real-world 3D scans of people in clothing, post-processed to be minimally noise-free 3D meshes (i.e., removing holes, self-intersection).
We train on 338 identities, each rescaled to a common scale in the unit bounding box as a simple data pre-processing step.
To learn the structured pose space, we used the Mixamo dataset~\cite{mixamo} and animated canonically-posed identities. 
Mixamo provides 3D human motions from which we collect a set of 2,446 motion sequences covering a wide variety of action categories of daily activities and sports. 
From this set of posed, clothed people, we randomly sample scan-motion pairs and obtain 40K randomly posed instances, without requiring seeing multiple identities in the same pose.
We evaluate our method on six unseen test identities performing various dancing moves, comprising to a total of 540 test frames organized in 90-frame sequences per identity.

\noindent\textbf{Evaluation metrics.}
To quantitatively evaluate model fitting to depth sequences, we measure reconstruction quality as well as tracking performance.
To measure reconstruction quality, we follow the evaluation protocol of OccupancyNets~\cite{mescheder2019occupancyNet} and compute Intersection over Union, Chamfer distance, and normal consistency on a per-frame basis. 
\mbox{\emph{Intersection over Union} (IoU)} measures overlap between the predicted and ground truth meshes, and is computed over $10^6$ randomly sampled points from the unit bounding box.
\mbox{\emph{Chamfer}-$\ell_2$ (C-$\ell_2$)}  measures the bi-directional distance between the prediction and ground truth with 100k randomly sampled points on the surfaces, giving distance characterization to any potentially mismatched surface reconstruction.
\mbox{\emph{Normal Consistency} (NC)} measures surface quality as the mean absolute dot product of the normals of the predicted mesh with the normals from the corresponding nearest neighbors in the ground truth mesh.
Finally to measure tracking performance, we follow the evaluation protocol of prior non-rigid tracking works~\cite{bozic2020neural,palafox2021npms} and evaluate \mbox{\emph{End-Point Error} (EPE)} as the average $\ell_2$ distance between predicted and ground truth deformations.
%

\subsection{Model Fitting to Monocular Depth Sequences}
\label{sec:model_fitting_experiments}

We evaluate our \OURS{} model fitting to new monocular depth sequences in comparison with state of the art on monocular depth sequences rendered from our Renderpeople~\cite{renderppl}-constructed dataset.
We compare with the state-of-the-art Neural Parametric Models (NPMs)~\cite{palafox2021npms} and \mbox{IP-Net~\cite{bhatnagar2020ipnet}}. 
We train NPMs on our Renderpeople training split.
Since NPMs require watertight meshes for training to determine inside/outside, and inside/outside queries on our train data tend to be unreliable in often-articulated regions such as hands, we adapt NPMs as NPMs* which incorporates the Eikonal loss, SIREN activations, and FiLM conditioning of our approach.
For \mbox{IP-Net~\cite{bhatnagar2020ipnet}}, we use a model checkpoint provided by the authors, which was also used to evaluate on RenderPeople.

\begin{figure*}
    \centering
    \fontsize{9pt}{11pt}\selectfont
    \def\svgwidth{0.95\linewidth}
    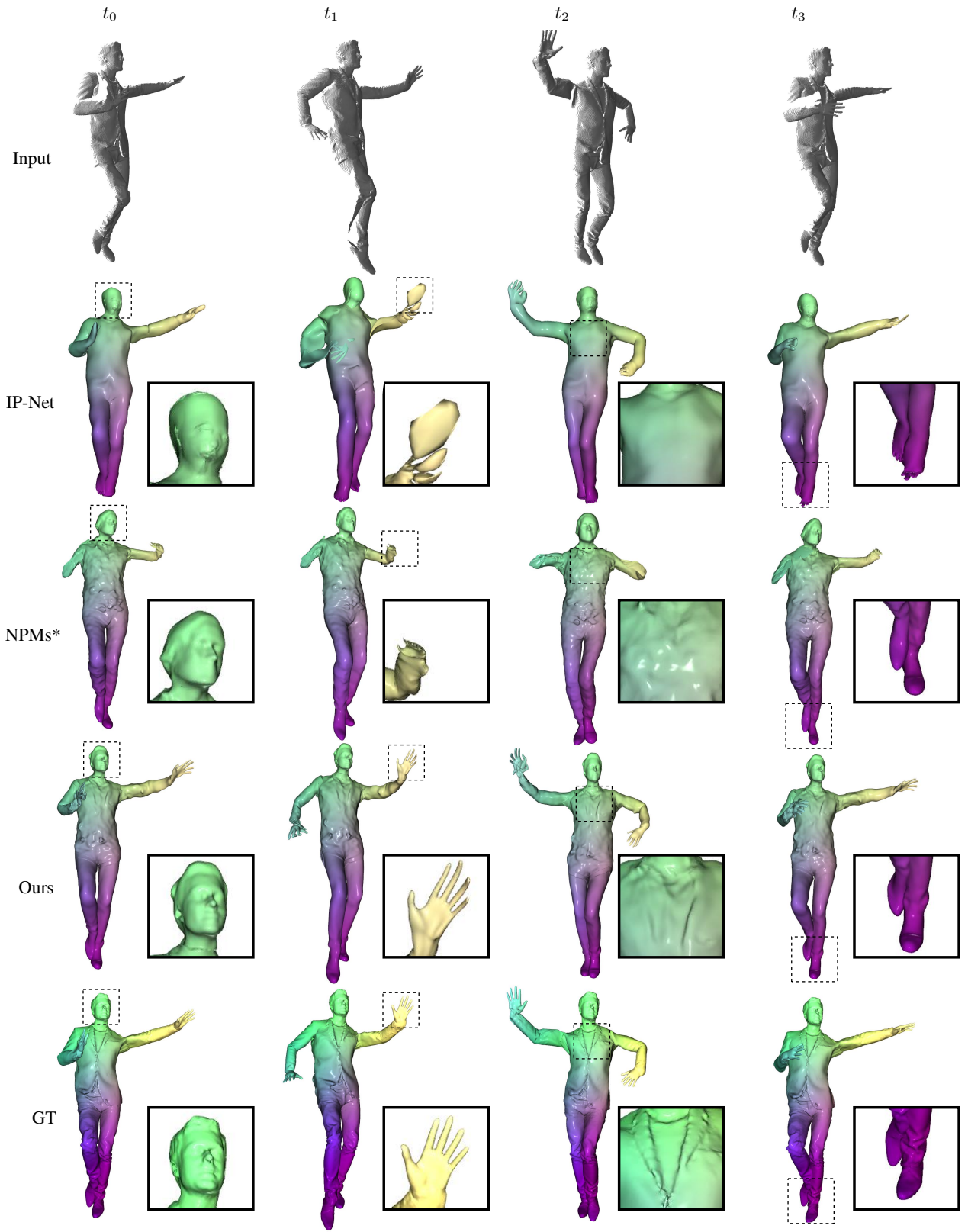
    \vspace{-0.2cm}
    \caption{
        Comparison to the state-of-the-art NPMs* \cite{palafox2021npms} and IPNet~\cite{bhatnagar2020ipnet} on the task of model fitting to a monocular depth sequence (first row).
        Our \OURS{} show superior tracking performance, as well as a better capability to capture high-frequency details (see jacket boundary on the third column).
    } 
    \label{fig:qualitative_comparison}
\end{figure*}

Table~\ref{tab:quantitative_comparison} shows a quantitative comparison with NPMs* and IP-Net on fitting to monocular depth sequences.
Our structural, part-driven representation of shape and pose produce notably improved reconstruction and tracking performance.
In particular, our \OUR{} achieves higher IoU and normal consistency, as well as significantly reduced Chamfer distance and 3D End-Point-Error, indicating more globally consistent tracking and reconstruction leveraging part-based fitting.
Qualitative comparisons are depicted in Figure~\ref{fig:qualitative_comparison}.
Under significant, complex motion in the input sequence, our \OUR{} maintains robust tracking and consistent geometry, while NPM* fails to capture more dramatic motions (e.g., in the arms).

\begin{table}[t]
    \centering
	\resizebox{0.75\linewidth}{!}
	{%
    \begin{tabular}{lcccc}
        \toprule
        \textbf{Method} & \textbf{IoU} $\uparrow$ & \textbf{C-$\ell_2$} $\downarrow$ & \textbf{NC} $\uparrow$ & \textbf{EPE} $\downarrow$ \\ 
        \midrule
        IP-Net           & 0.729 & 0.00053 & 0.837 & 0.168 \\
        NPMs*           & 0.755 & 0.00163 & 0.856 & 0.053 \\
        {\bf Ours} & \textbf{0.785} & \textbf{0.00032} & \textbf{0.883} & \textbf{0.034} \\
        \bottomrule
    \end{tabular}
    }
    \vspace{-0.2cm}
    \caption{
        Comparison with state-of-the-art NPMs* on the test set of our RenderPeople dataset.
        Our part-aware decomposition enables notably improved reconstruction and tracking.
        }
    \label{tab:quantitative_comparison}
\end{table}

\subsection{Ablations}
\label{subsec:ablations}

\noindent\textbf{Robustness to Pose Code Initialization.}
We observe that our part-aware disentanglement of shape and pose provides significant robustness in pose tracking, and in particular, maintains robustness in the absence of pose encoder initialization (which may not be available in scenarios such as generalizing to different sensor inputs).
We demonstrate this in Table~\ref{tab:ablations}, comparing with a variant of our approach without pose encoder initialization (w/o PE) and instead using random initialization for pose codes.
This results in very poor initial pose estimates, with an effectively random initial set of deformation fields.
While this poor initialization results in slightly impaired performance, our structurally-guided \OURS{} nonetheless can recover a significant portion of the reconstruction and tracking performance of using the pose encoder initialization.
In contrast, we observe that the full-shape shape and pose encoding of NPMs* fails to recover from poor pose initializations.
We refer to the supplementary material for further qualitative visualizations.

\begin{table}[t]
	\resizebox{\linewidth}{!}{%
    \centering
    \begin{tabular}{lcccc}
        \toprule
        \textbf{Method} & \textbf{IoU} $\uparrow$ & \textbf{C-$\ell_2$} $\downarrow$ & \textbf{NC} $\uparrow$ & \textbf{EPE} $\downarrow$ \\ 
        \midrule
        NPMs* (w/o PE) & 0.269 & 0.01397 & 0.658 & 0.182 \\
        \midrule
        Ours (w/o PE, w/o PGM)  & 0.671 & 0.00070 & 0.836 & 0.052 \\
        Ours (w/o PE)           & 0.681 & 0.00065 & 0.839 & 0.052 \\
        Ours (w/o PGM)          & 0.766 & 0.00037 & 0.874 & 0.037 \\
        \midrule
        {\bf Ours} & \textbf{0.785} & \textbf{0.00032} & \textbf{0.883} & \textbf{0.034} \\
        \midrule
        {\bf Ours (w/ GT PS)}            & \textbf{0.809} & \textbf{0.00021} & \textbf{0.894} & \textbf{0.026} \\
        \bottomrule
    \end{tabular}
    }
    \vspace{-0.2cm}
    \caption{
        Ablation studies.
        We evaluate the effects of a learned pose encoder initialization (PE), using our part grid mask (PGM) in test-time fitting, and we additionally study the impact of ground truth part segmentation (GT PS) of input depth maps.
        Our part-driven shape and pose spaces provide significant robustness against lack of pose initializations, and leveraging part mask guidance for optimization additionally improves performance.
        }
    \label{tab:ablations}
\end{table}

\noindent\textbf{Part-based Grid Masking.}
Table~\ref{tab:ablations} additionally evaluates the effect of leveraging part information as a part grid mask (PGM) during test-time optimization; this local part focus enables more accurate reconstruction and tracking.

\noindent\textbf{Limitations.}
Our structured parametric modeling of deformable objects enables  robust model fitting to challenging monocular depth sequences, but maintains several limitations.
For instance, our local pose spaces do not characterize potential high-level motion priors given by motion very far from the local region (e.g., one hand moving back more often occurs with the other hand moving forward than not), which could provide additional global context.
Additionally, very fine-scale sharp details can become oversmoothed across the global optimization, which could potentially be characterized with perceptually-oriented measures.
\vspace{0.35cm}
\section{Conclusion}
In this work we have introduced \OURS{}, a deformable object representation where non-rigid object motion is structurally decomposed into part-based disentangled representations of shape and pose.
Our structured characterization of object movement can be leveraged at test time to fit to input depth sequences of unseen shapes, leveraging part-based correspondences to establish robust optimization to fit to the input sequence.
Our experiments show significantly improved robustness in reconstruction and tracking, particularly in scenarios of challenging, complex motions in observed depth sequences.
We believe that this representation will be useful in a variety of spatio-temporal tasks.

\section*{Acknowledgments}
This work was primarily done during an internship at Meta Reality Labs Research. We would additionally like to thank Yuanlu Xu for informative discussions, and support from the Bavarian State Ministry of Science and the Arts and coordinated by the Bavarian Research Institute for Digital Transformation (bidt).

\begin{appendix}
\section*{Appendix}
%
%
In this appendix, we provide additional details for our test-time optimization in Sec.~\ref{sec:testtime}, and then present an ablation study on the use of our pose encoder for pose code initialization in Sec.~\ref{sec:pose_code_init}.
In Sec.~\ref{sec:additional_comp} we present a quantitative comparison with IP-Net \cite{bhatnagar2020ipnet}.
Additional qualitative evaluations and results are shown in the supplemental video.
 
\section{Test-time Optimization}
\label{sec:testtime}

In Eq.~6 in the main paper we present the energy term that is minimized at test-time when fitting our \OURS{} to a depth sequence, which we rewrite here for completeness:
\begin{equation}\small
    [\tilde{\fs}_{\rqq}], \{ [\tilde{\fp}_{\rqq}^j] \}_{j=1}^L = \argmin_{[\fs_{\rqq}], \{ [\fp_{\rqq}^j] \}_{j=1}^L} \sum_{j=1}^L \sum_{\forall \fx_k} \mathcal{L}_{\mathrm{r}} + \mathcal{L}_{\mathrm{c}} + \mathcal{L}_{\mathrm{t}} + \mathcal{L}_{\mathrm{icp}}.
    \label{eq:test_time_optim}
\end{equation}
As mentioned in the paper, $\mathcal{L}_{\mathrm{c}}$ enforces shape and pose code regularization through an $\ell_2$ loss on the latent codes:
\begin{equation}\small
    \mathcal{L}_{\mathrm{c}} = \sum_{q=1}^{Q} \frac{\norm{\fs_{\rqq}}^2_2}{\sigma_{\rs}^2} + \frac{\norm{\fp_{\rqq}^j}^2_2}{\sigma_{\rp}^2},
\end{equation}
with $\sigma_{\rs}^2 = 0.01$, $\sigma_{\rp}^2 = 0.001$.

$\mathcal{L}_{\mathrm{t}}$ enforces temporal regularization between the current frame $j$ and its neighboring frames \mbox{$H = \{ j-1, j+1 \}$}.
As in \cite{palafox2021npms}, this is enforced with an $\ell_2$-loss on the pose MLP flow predictions for points $\fx_k$, and controlled with a weight of $\lambda_t = 10$:
\begin{equation}
    \mathcal{L}_{t} = \lambda_t \sum_{q=1}^{Q} \sum_{h \in H} \norm{ f_{\theta_{\rp}^{\rqq}}(\fs_{\rqq}, \fp_{\rqq}^j, \fx_k) - f_{\theta_{\rp}^{\rqq}}(\fs_{\rqq}, \fp_{\rqq}^h, \fx_k) }^2_2.
\end{equation}

As presented in the Sec.~3.5.1 in the main paper, we employ a part-guided ICP-inspired loss $\mathcal{L}_{\mathrm{icp}}$.
We recall that $\mathcal{L}_{\mathrm{icp}}$ is computed by establishing part-driven correspondences between each estimated part $q$ from the predicted input depth map part labels and the part decoder predictions.
To this end, every $I_{\rm{resample}}$ iterations we consider the canonically-posed shape from the current state of the shape codes, re-sample a new set of \mbox{$N_t = \SI{500}{}$k} points around the mesh and keep those within a distance $\epsilon_{\mathrm{icp}} = 0.005$ (in normalized units), denoted by $\fx_k^{\mathrm{ns}}$, from the implicitly represented surface.
We use our part decoder (Sec.~3.5 in the main paper) to estimate part labels for these canonical points, and then warp them into a posed frame $j$ using our pose decoders.
Then for every point in the input depth map $\bm{x}' \in D_j$, and given its predicted part $q$ (obtained by PointNet++), we find its nearest neighbor in the warped set of points belonging to $q$ based on $f_{\theta_{\rqq}}$, denoted by
\begin{equation}
\mathcal{W}_{\rqq} = \{ \fx_k^{\mathrm{ns}} + f_{\theta_{\rp}^{\rqq}}(\fs_{\rqq}, \fp_{\rqq}^j, \fx_k^{\mathrm{ns}}) \}
\end{equation}
to establish correspondences, and minimize the distance between these points:
\begin{equation}
    \mathcal{L}_{\mathrm{icp}} = \lambda_{\mathrm{icp}} \sum_{q=1}^{Q} \sum_{\bm{x}' \in D_j} \norm{ \bm{x}' - \mathrm{NN}_{\mathcal{W}_{\rqq}}(\bm{x}') }_2.
    \label{eq:icp_term}
\end{equation}
In the above equation, $\mathrm{NN}_{\mathcal{W}_{\rqq}}(\cdot)$ denotes a function that queries the nearest neighbor of a 3D point in a set of points $\mathcal{W}_{\rqq}$.
We control the importance of this loss with $\lambda_{\mathrm{icp}} = 20$ in our experiments.

Finally, we control $\mathcal{L}_{\mathrm{r}}$ (Eq.~7 in the main paper) with $\lambda_{\mathrm{r}} = 1$.

\vspace{0.5cm}
Optimizing over an input sequence of 90 frames until convergence (for 200 optimization steps) takes approximately 1.5 hours on a GeForce~RTX~3090 with our highly unoptimized implementation.

\subsection{Effect of Pose Code Initialization}
\label{sec:pose_code_init}
\begin{figure*}
    \centering
    \fontsize{9pt}{11pt}\selectfont
    \def\svgwidth{0.9\linewidth}
    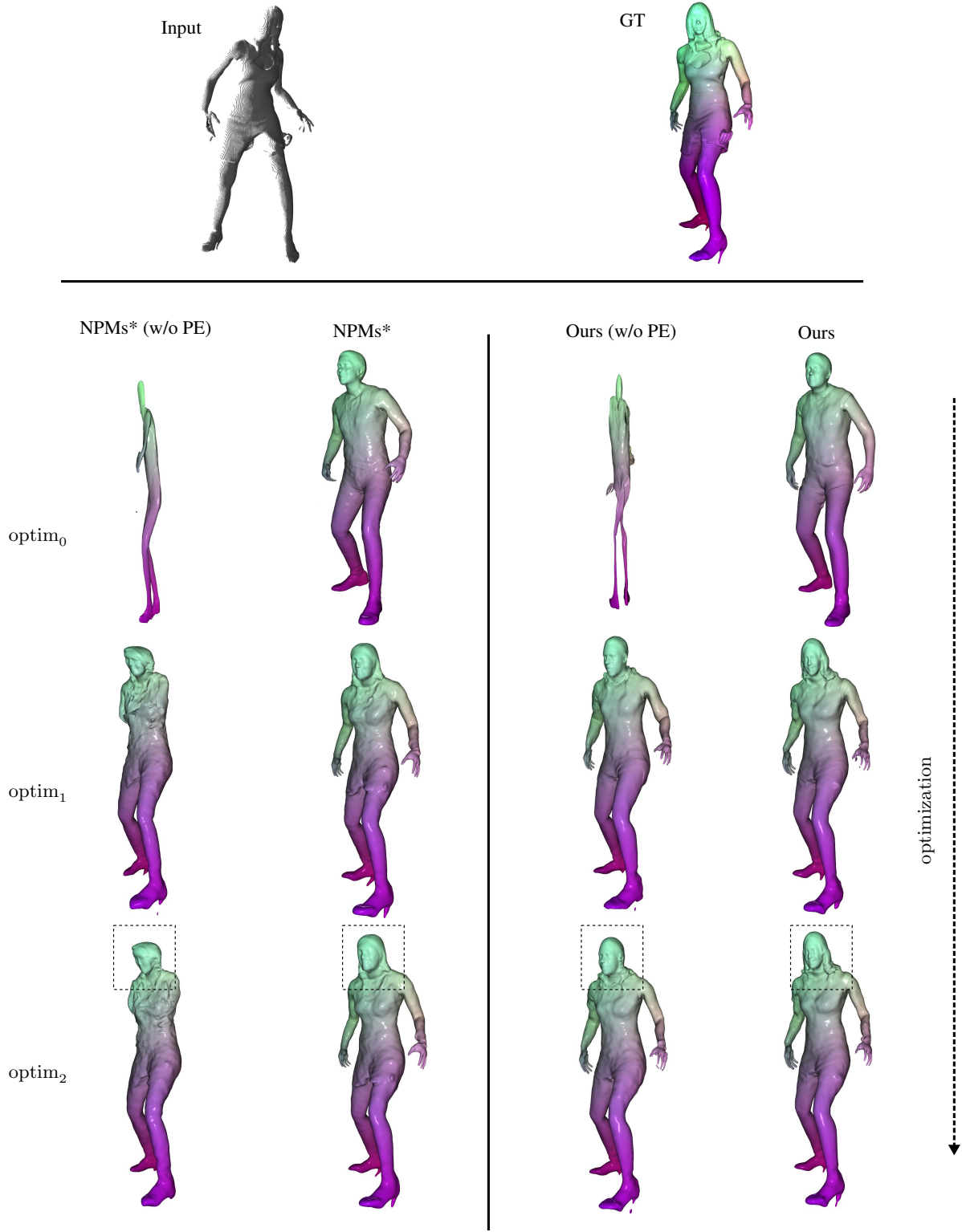
    \caption{
        For a given frame, we study how optimization evolves across different optimization steps for NPMs* (with and without pose encoder initialization) and our \OURS{} (with and without pose encoder initialization).
        Note that \OURS{} are robust to pose code initialization, and can recover tracking even when starting from randomly initialized pose codes.
    } 
    \label{fig:pose_enc_comp}
\end{figure*}

We study the effect of pose code initialization in Fig.~\ref{fig:pose_enc_comp}.
For a given frame, we study how optimization evolves across different optimization steps for NPMs* \cite{palafox2021npms} (with and without pose encoder initialization) and our \OURS{} (with and without pose encoder initialization).
Our part basis helps to establish global correspondences that provide robustness against lack of good pose initialization.

\section{Additional Comparisons to State of the Art}
\label{sec:additional_comp}

\begin{figure*}
    \centering
    \fontsize{9pt}{11pt}\selectfont
    \def\svgwidth{0.95\linewidth}
    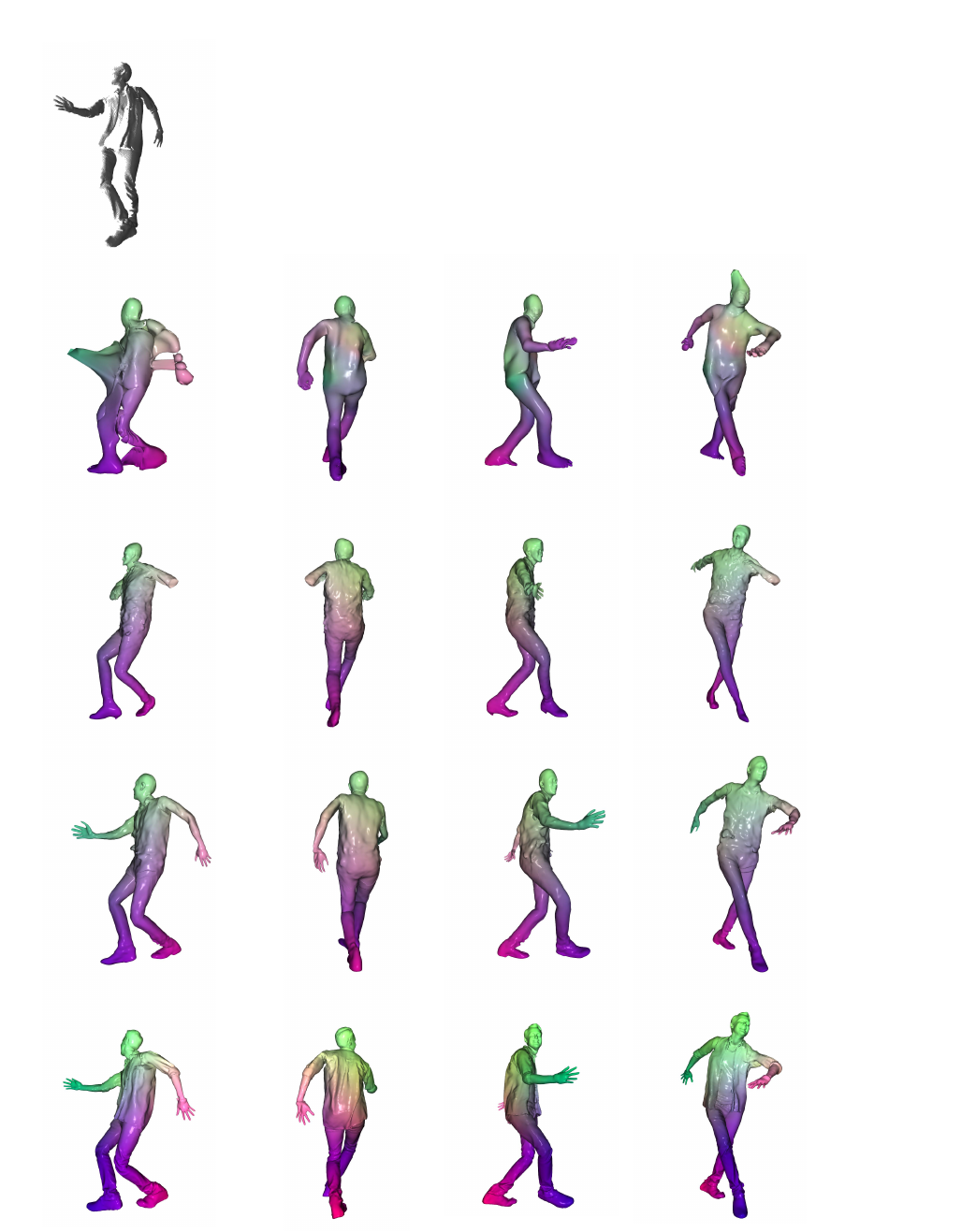
    \caption{
        Qualitative comparison to NPMs* \cite{palafox2021npms} and IP-Net \cite{bhatnagar2020ipnet} on the task of model fitting to a monocular depth sequence.
        In complex motion scenarios such as loop closures, NPMs* and IP-Net struggle to track the motion, whereas our \OURS{} robustly maintains tracking.
    } 
    \label{fig:qualitative_comparison_suppl}
\end{figure*}

In Fig.~\ref{fig:qualitative_comparison_suppl} we show a qualitative comparison with IP-Net and NPMs* on one of our test sequences; we show superior performance in loop closing, demonstrating our tracking robustness while maintaining detailed geometry.

\end{appendix}

\clearpage
{\small
\bibliographystyle{ieee_fullname}
\bibliography{refs}
}

\end{document}